\documentclass{article}
\usepackage{ijcai24}

\usepackage{times}
\usepackage{soul}
\usepackage{url}
\usepackage[hidelinks]{hyperref}
\usepackage[utf8]{inputenc}
\usepackage[small]{caption}
\usepackage{graphicx}
\usepackage{amsmath}
\usepackage{amsthm}
\usepackage{amssymb}
\usepackage{booktabs}
\usepackage{algorithm}
\usepackage{algpseudocode}
\usepackage[switch]{lineno}
\usepackage{array}
\DeclareMathOperator*{\argmax}{arg\,max}

\usepackage{multirow}
\usepackage[capitalise]{cleveref}

\title{Bias Mitigation in Fine-tuning Pre-trained Models for Enhanced Fairness and Efficiency}

\author{
Yixuan Zhang$^1$
\and
Feng Zhou$^2$\and
\affiliations
$^1$Hangzhou Dianzi University\\
$^2$Remin University of China\\
\emails
yixuan.zhang@hdu.edu.cn,
feng.zhou@ruc.edu.cn
}

\begin{document}

\maketitle

\begin{abstract}
Fine-tuning pre-trained models is a widely employed technique in numerous real-world applications. However, fine-tuning these models on new tasks can lead to unfair outcomes. This is due to the absence of generalization guarantees for fairness properties, regardless of whether the original pre-trained model was developed with fairness considerations. To tackle this issue, we introduce an efficient and robust fine-tuning framework specifically designed to mitigate biases in new tasks. Our empirical analysis shows that the parameters in the pre-trained model that affect predictions for different demographic groups are different, so based on this observation, we employ a transfer learning strategy that neutralizes the importance of these influential weights, determined using Fisher information across demographic groups. Additionally, we integrate this weight importance neutralization strategy with a matrix factorization technique, which provides a low-rank approximation of the weight matrix using fewer parameters, reducing the computational demands. Experiments on multiple pre-trained models and new tasks demonstrate the effectiveness of our method. 
\end{abstract}

\section{Introduction}
Over recent decades, automated decision-making systems have been applied in extensive application in numerous fields, including medicine~\cite{medical}, finance~\cite{credit_example}, criminology~\cite{compas_juve}, etc. The broad adoption of machine learning techniques raises concerns regarding its potential to exhibit unfair behavior, as these systems are data-driven and can inherit the biases present in their input data. This replication of bias by the models can result in biased decisions that unfairly discriminate, posing significant risks to individuals and society~\cite{bird2016exploring}. Consequently, it is crucial to develop machine learning algorithms that are free from discrimination against specific demographic groups. 

Existing literature indicates two primary strategies for achieving fairness in machine learning models. The first approach involves integrating fairness constraints directly into the training phase, often referred to as in-processing. This method modifies the objective functions to account for fairness considerations. Various studies have explored this avenue by incorporating fairness constraints into the optimization process, aiming to mitigate biases in the learned model~\cite{2016_zafar,2018_icml_reductions,Kamishima_2011}.
In contrast, the second strategy involves learning fair representations as a pre-processing step. This method focuses on creating unbiased and fair representations of the input data before employing conventional machine learning techniques. These fair representations serve as the foundation for subsequent learning tasks, ensuring that the underlying data used for training the models is inherently unbiased~\cite{icml_2013,NIPS2017_optimised_preprocessing,pmlr-v206-zhang23g}. 

The preceding methods, whether in-processing or pre-processing, require constructing a new model from scratch for a specific task. However, in practical scenarios, retraining a model from the ground up for every new task is computationally intensive. Therefore, a more common approach is to fine-tune existing pre-trained models. For instance, with the prevalence of large models today, when adapting to new task domains, it is impractical to train these massive models from scratch. Instead, we often opt to fine-tune them by, for example, freezing the feature extraction part and only training the final linear layer. 
However, fine-tuning pre-trained models on new tasks can result in unpredictable unfair outcomes. This unpredictability remains even when the pre-trained model was trained with fairness objectives, primarily because there is a lack of assurance regarding the fairness property's generalizability~\cite{Kamishima_2011,NEURIPS2020_mmd}. 

We employ a transfer learning strategy to address the specified limitations, focusing on neutralizing the importance of influential weights to reduce bias. 
The intuition is clear: from our experiments, it is evident that distinct weights in the pre-trained model have a substantial impact on predictions across different demographic groups. Therefore, our objective is to balance the impact of these influential weights, making it challenging to differentiate group information from the weights, thus mitigating bias. 
Due to the enormous parameter count in large pre-trained models, even fine-tuning just the linear layer requires a considerable amount of time. Therefore, to further improve efficiency, we enhance the method by approximating the weights in the linear layer with fewer parameters using singular value decomposition (SVD)~\cite{svd}. 
Our method consists of three steps: 
(1) We assess the importance of linear-layer weights by employing Fisher information across various demographic groups, subsequently neutralizing these importance scores. (2) Utilizing these neutralized importance scores, we conduct a weighted SVD of the linear-layer weight matrix. (3) We replace the original linear layer with the low-rank layers obtained from SVD and fine-tune them for the new task. 

Specifically, we make the following contributions: 
(1) We propose a weight importance neutralization strategy to mitigate bias for fine-tuning on new tasks, taking into account the influence of each parameter on prediction across different demographic groups. 
(2) To enhance the fine-tuning efficiency, we use SVD to create a low-rank approximation of the linear-layer weight matrix, which has fewer parameters. 
(3) Our empirical analysis shows that even with a fair pre-trained model, fine-tuning on new tasks can result in severe bias, and our method can effectively address this issue. 

The structure of this paper is as follows: \cref{sec: related_work} reviews prior work in the domain of fairness-aware learning, model fine-tuning, and low-rank approximation. \cref{sec: method} details the methodology we propose. In \cref{sec: experiment}, we conduct the experimental comparison of our method with various baselines and leading-edge techniques across three real-world datasets. Lastly, \cref{sec: conclusion} summarizes this work and outlines potential avenues for future research. 

\section{Related Work}
\label{sec: related_work}
In this section, we briefly review fairness-aware learning methods, model fine-tuning, and low-rank approximation literature which are most relevant to ours. 

\subsection{Bias Mitigation}
Numerous recent studies have demonstrated that modern machine learning models exhibit biases against certain demographic groups~\cite{pmlr-v81-buolamwini18a}. Existing bias mitigation methods can be generally grouped into two categories. The first focuses on the in-processing stage to remove discrimination, such as modifying the objective functions~\cite{pmlr-v119-roh20a,2018_icml_reductions}, employing adversarial learning~\cite{zhang_adversarial} or using boosting techniques~\cite{adafair}. This kind of approach aims for a fair classifier that ensures predictions are independent of sensitive variables. Typically, the methods in this category require annotations of sensitive variables so that the fairness criteria can be applied in the training process. 

The second family of bias mitigation methods is based on representation learning. It involves developing different neural network architectures with modified learning strategies to learn representations that remove the undesirable relationship between sensitive variables and non-sensitive variables in the raw data~\cite{DBLP:flex,2011_dwork,icml_2013}. In this setting, the learned representations preserve information that is useful for prediction while removing the dependency on sensitive variables.

\subsection{Adaptation and Fine-tuning}
Adaptation and fine-tuning from pre-trained models is a widely used technique in various real-world applications. 
Previously, comprehensive works used full fine-tuning, which involves initializing the model with pre-trained parameters across all layers, and then subsequently training the model on specific downstream tasks~\cite{liu2019roberta,tayaranian-hosseini-etal-2023-towards}. However, with the growth in the number of parameters of deep learning models (with billions of parameters), it becomes challenging to adapt these models to downstream tasks, therefore, parameter-efficient fine-tuning has been proposed to reduce the computational demands, which introduces new, trainable parameters for task-specific fine-tuning~\cite{he2022towards,lin-etal-2020-exploring}. Beyond the methods mentioned above, partial fine-tuning is another strategy, focusing on updating only a selected subgroup of pre-trained parameters that are crucial for downstream tasks~\cite{xu-etal-2021-raise,10.1609/aaai.v37i11.26505}. The recent popular methods in the low-rank decomposition family are categorized as reparameterized fine-tuning, which utilizes low-rank decomposition to reduce the number of trainable parameters, particularly useful for pre-trained weights~\cite{hu2022lora,aghajanyan-etal-2021-intrinsic}. In this work, we adopt a hybrid fine-tuning strategy, combining the benefits of partial fine-tuning with the low-rank decomposition method.

\subsection{Low-rank Approximation}
Fine-tuning solely the linear layer of large pre-trained models can still demand significant computational resources due to their extensive parameter count. To improve efficiency during the fine-tuning phase, strategies like low-rank decomposition have been introduced. These techniques aim to closely approximate the weight matrix in the linear layer with a more compact version~\cite{hu2022lora,jaderberg2014speeding}. 
Recently, SVD has been widely applied in low-rank approximation. This approach is geared towards compressing the weight parameters to improve efficiency, as noted in several studies~\cite{ben-noach-goldberg-2020-compressing,acharya2018online}. By employing SVD and similar techniques, the goal is to significantly minimize the reconstruction error of the weight matrix by using fewer parameters, thereby achieving a more efficient model without substantially sacrificing performance.

\section{Method}
\label{sec: method}
In this section, we first introduce the notations used in \cref{sec:method_notation}. Then, we present empirical findings that illustrate how fine-tuning a pre-trained model on new tasks can lead to unfair results, as detailed in \cref{sec:method_analysis}. After this analysis, we introduce our weight importance neutralization technique in \cref{sec:method_weight_neutralization}, specifically formulated to facilitate fair fine-tuning on new tasks.

\subsection{Notations}
\label{sec:method_notation}
We focus on the supervised learning task in the context of binary classification. The training set of a new task is defined as $\mathcal{D} = \{(x_i, y_i, s_i)\}_{i=1}^{N}$, where $x_i\in\mathcal{X}\subset\mathbb{R}^D$ denotes the high-dimensional non-sensitive feature, $y_i\in\mathcal{Y}=\{-1,1\}$ denotes the binary ground-truth label, and $s_i \in\mathcal{S}=\{1,2\}$ denotes the binary sensitive feature, such as gender or race. 
$\mathcal{D}_{S=s} = \{(x_i, y_i, s_i=s)\}_{i=1}^{N_s}$ denotes a subset of the dataset, which consists of samples sharing the same sensitive feature. 
The pre-trained model is expressed as $f(x;\theta)$ with $\theta$ parameterizing the model. The pre-trained model $f(x;\theta)$ can generally be divided into two sequential parts: a feature extractor that transforms the raw data into a representation, and a classification head (usually a linear layer) that converts the representation into output probabilities. The predicted class is determined by $\hat{y}=\argmax f(x;\theta)$. The pre-trained model is fine-tuned on the dataset $\mathcal{D}$ to adapt to a new task. In this process, the feature extractor is frozen, and only the classification head is fine-tuned. Our primary objective is to alleviate discrimination during fine-tuning.

\subsection{Fine-tuning Introduces Bias}
\label{sec:method_analysis}
This section starts with an exploration of the discrimination caused by fine-tuning the pre-trained models on new tasks. To this end, we pre-train an MLP on the Adult Income dataset~\cite{adultcensus}, using 60\% of the data for pre-training and reserving the rest 40\% as the new task. Due to the inherent presence of discrimination in the Adult Income dataset, the pre-trained model obtained in this way is considered unfair, denoted as $f_B$. As a control group, we also pre-train an MLP with the same data but using the in-processing fairness-aware learning approach~\cite{fair_constraints}. In this approach, we incorporate a demographic parity constraint into the objective function. We treat this pre-trained model that considers fairness as a fair one, which is denoted as $f_F$. Then we fine-tune both $f_F$ and $f_B$ on the new task.

\begin{figure}[t]
    \centering
    \includegraphics[width=0.9\linewidth]{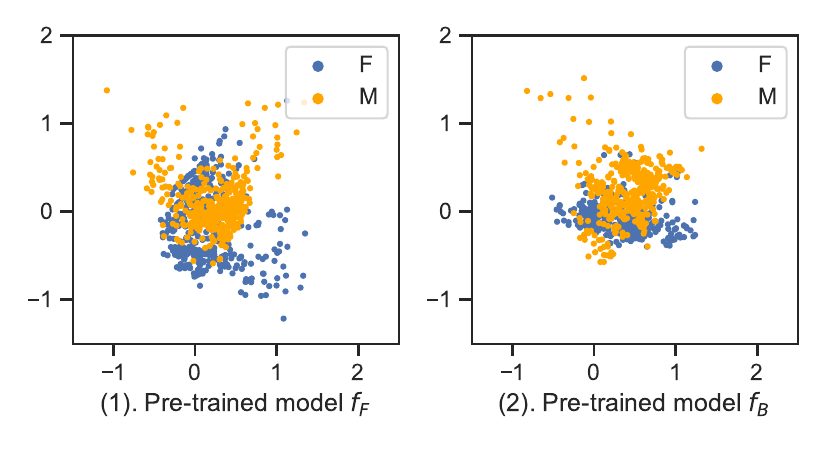}
    \caption{Principal component analysis on the representations from the linear layer after fine-tuning the pre-trained models. Blue points represent the ``Female" group, while orange points represent the ``Male" group.}
    \label{fig:rep}
\end{figure}

For ease of visualization, we apply the principal component analysis (PCA) to the representation obtained from $f_F$ and $f_B$ after fine-tuning, reducing it to 2 dimensions. 
\cref{fig:rep} illustrates the distribution of the representations after dimensionality reduction for samples predicted as positive in the ``Male" and ``Female" groups for $f_F$ and $f_B$. 
It is evident that the representations obtained from $f_B$ display a different pattern across different demographic groups. While for $f_F$, the pattern is less pronounced than that of $f_B$, yet there is still a noticeable division into two groups in the plot. 

These findings demonstrate that bias can be introduced when a pre-trained model is fine-tuned on a new dataset. This phenomenon occurs regardless of whether the original pre-trained model is developed with fairness considerations or not. 
This discovery is significant, as it suggests that many large language models (LLM) obtained from fine-tuning may also encounter fairness issues, such as ChatGPT fine-tuned from GPT-3 and Sparrow fine-tuned from Chinchilla. 

\begin{figure*}[t]
    \centering
    \includegraphics[width=1\linewidth]{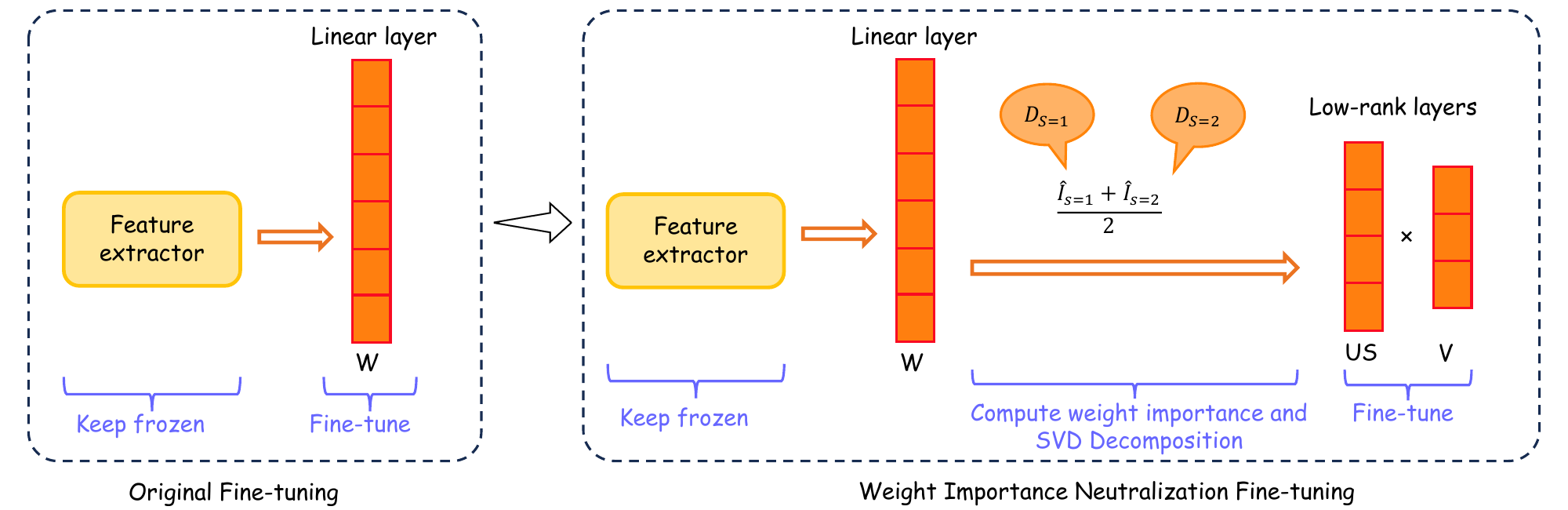}
    \caption{The weight importance neutralization fine-tuning. \textbf{Left:} the original fine-tuning method, \textbf{Right:} our proposed fine-tuning method.}
    \label{fig:overview}
\end{figure*}





\subsection{Weight Importance Neutralization}
\label{sec:method_weight_neutralization}
Based on the empirical findings in \cref{sec:method_analysis}, this section presents our \textit{Weight Importance Neutralization} method. This approach effectively reduces the bias inherited from fine-tuning the pre-trained model on new tasks. We first address fundamental questions: What is weight importance? Why do we consider it as the starting point of our method?

\begin{figure}[ht]
    \centering
    \includegraphics[width=1\linewidth]{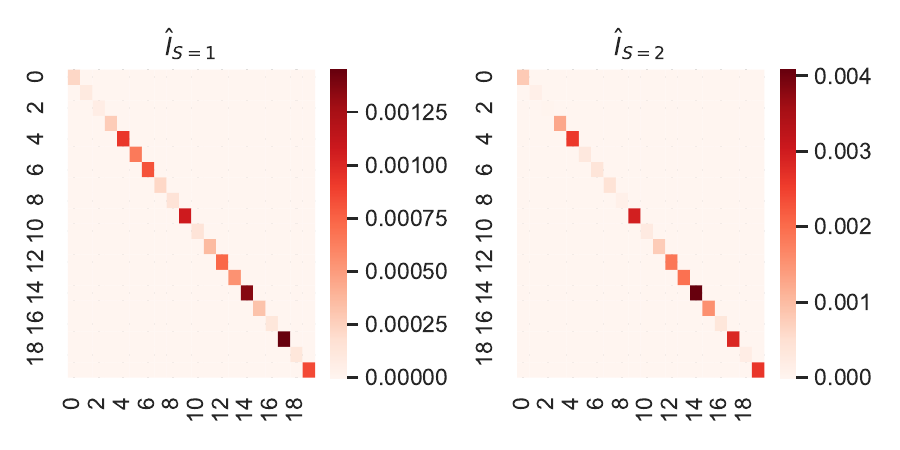}
    \caption{The diagonal entries of the Fisher information matrix for all parameters of the final linear layer of the model. \textbf{Left}: the model is trained on $\mathcal{D}_{S=\text{``Female"}}$, \textbf{Right}: the model is trained on $\mathcal{D}_{S=\text{``Male"}}$.}
    \label{fig:fisher_matrix_heatmap}
\end{figure}

\textbf{Weight Importance via Fisher Information.} The concept of weight importance is derived from Fisher information~\cite{fisher_info}, which is a metric used to evaluate how crucial certain parameters are to the model's prediction. Similar to \cref{sec:method_analysis}, we conduct experiments on the Adult Income dataset with an MLP. As illustrated in \cref{fig:fisher_matrix_heatmap}, we plot the diagonal entries of the Fisher information matrix for all parameters (20 in total) of the final linear layer of the model. This is done separately for the model trained by $\mathcal{D}_{S=1}$ (represents ``Female" group) or $\mathcal{D}_{S=2}$ (represents ``Male" group). From \cref{fig:fisher_matrix_heatmap}, it is clear that the influential weights for prediction across two demographic groups vary. Taking inspiration from this observation, we employ Fisher information to quantify the impact of parameters on the performance of the model trained by $\mathcal{D}_{S=1}$ or $\mathcal{D}_{S=2}$, and our goal is to mitigate this difference by neutralizing it. 

In practice, the exact form of Fisher information is normally intractable due to the marginalization over the space of data, so we use the empirical version to approximate the exact form. The empirical version is given by: 
\begin{equation}
\begin{aligned}
\label{eq:fisher_information}
    I_\theta &= \mathbb{E}\left [ \left ( \frac{\partial}{\partial \theta} \log p(\mathcal{D}\mid \theta) \right )^2 \right]\\
    & \approx \frac{1}{N}\sum^{N}_{i=1}\left(\frac{\partial}{\partial \theta} \mathcal{L}(x_i,s_i,y_i;\theta)\right)^2 =\hat{I}_\theta, 
\end{aligned}
\end{equation}
where $\mathcal{L}$ is the objective function of the corresponding model.



\textbf{Weight Importance Neutralization.} Because incorporating the full Fisher information matrix into our weighting strategy would be computationally intensive, we follow the same assumption setting in \cite{hsu2022language} where each row of the weight matrix is assumed to share the same importance score. 
Specifically, we explicitly denote the weight and bias parameters from $\theta$ as $W$ and $b$, respectively. Instead of calculating the entire matrix, we define the estimated Fisher information as $\hat{I}_{W_i} = \sum_j \hat{I}_{W_{ij}}$. Subsequently, we assume $\hat{I} = \text{diag}(\sqrt{\hat{I}_{W_1}},\cdots,\sqrt{\hat{I}_{W_d}})$. 
This allows us to more effectively calculate the importance of each weight entry in relation to different tasks, denoted as $\hat{I}_{S=1}$ and $\hat{I}_{S=2}$ for $\mathcal{D}_{S=1}$ and $\mathcal{D}_{S=2}$, respectively. 
Finally, the neutralized Fisher information is expressed as: 
\begin{equation}
\label{eq: weight_neutralization}
    \hat{I}_N = \frac{1}{2}(\hat{I}_{S=1} + \hat{I}_{S=2}). 
\end{equation}

\textbf{Efficiency Boost via Low-rank Approximation.} In our method, we specifically target the linear layer for fine-tuning to improve the efficiency of adaptation from the pre-trained model to the downstream task. However, even with this approach, fine-tuning large pre-trained models can still be time-consuming. To address this issue, prior research has focused on freezing the weights of the pre-trained model and training specific layers indirectly by optimizing the rank-decomposition matrices that capture the changes in these layers during fine-tuning~\cite{hu2022lora}. 
This strategy is based on the assumption that the weight updates during the adaptation process exhibit a low ``intrinsic rank". Specifically, for a pre-trained weight matrix $W_0 \in \mathbb{R}^{d\times k}$, the update is formulated through a low-rank decomposition: $W_0 + \Delta W = W_0 + BA$, where $B \in \mathbb{R}^{d\times r}, A\in \mathbb{R}^{r\times k}$, and the rank $r$ is much smaller than both $d$ and $k$. 

Continuing this thread, the weights from the pre-trained model can be effectively approximated using standard factorization methods, with SVD being the most commonly employed. In this work, we also focus on this particular factorization approach. Through SVD, the weight matrix is factorized into three matrices, denoted as $U$, $S$, and $V$. Specifically, assuming the final linear layer with weight matrix $W \in \mathbb{R}^{d\times k}$ and bias $b\in \mathbb{R}^{1\times k}$, the final linear layer can be approximated via SVD as: 
\begin{equation}
    Z = XW + b \approx (XUS)V^{\top}+b, 
\end{equation}
where $X$ is the input of the linear layer. 

By decomposing $W$ into $USV^{\top}$, we can replicate the linear layer with two smaller linear layers, where the first layer employs a weight matrix of $US$, and the second uses the weight matrix $V$ and bias $b$. This approach effectively reduces the total number of parameters in a full-rank scenario by $d\times k - (d_r+k_r)$, where $d_r$ and $k_r$ represent the parameter counts in the first and second layers, respectively. Through this decomposition, the model can be simplified and the fine-tuning process can be accelerated. 

Direct SVD of the linear layer's weight matrix ensures efficiency and information retention, but risks inheriting original biases. Therefore, to guarantee fairness, we integrate the weight importance with the SVD. Specifically, we modify the SVD optimization objective $\Vert W-AB\Vert_2$ to include the neutralized Fisher information and obtain a weighted SVD: 
%

\begin{equation}
\label{eq: weighted_svd}
    \underset{A,B}{\text{min}}||\hat{I}_N W-\hat{I}_N AB||_2.
\end{equation}
where $A = US$, $B=V^{\top}$. Using standard SVD on $\hat{I}_N W$, we get $U^*, S^*$ and $V^*$, then the solution is computed by removing $\hat{I}_N$ from the factorized matrices, and the solution of $A$ will be $\hat{I}_N^{-1}U^*S^*$, and the solution of $B$ is ${V^*}^{\top}$. Finally, the weighted compressed $W$ can be rewritten as $A = \hat{I}_N^{-1}U^*_rS^*_r$ and $B = {V^*_r}^{\top}$, where $r$ indicates truncating $U^*$, $S^*$, and $V^*$ to preserve only the top $r$ ranks. 

The motivation behind our method is straightforward: by applying the neutralized importance score computed by Fisher information, we enable the decomposed matrix to recover the weights that are crucial for predicting both $\mathcal{D}_{S=1}$ and $\mathcal{D}_{S=2}$. Weights that are less important for predictions in either group are ignored. This can help us ensure the model's fairness by prioritizing weights important for predicting both demographic groups. An overview of our method is provided in \cref{fig:overview} and the overall process of our proposed weight importance neutralization fine-tuning is given in \cref{alg:algorithm}. 

\begin{algorithm}[t]
  \caption{Weight Importance Neutralization Framework}\label{alg:algorithm}
  \begin{algorithmic}[1]
   \State \textbf{Input:} Downstream task $\mathcal{D}_{S=1}$ and $\mathcal{D}_{S=2}$, pre-trained model $f$, loss function $\mathcal{L}$, learning rate $\eta$, training epochs $T$.
   \State $f' \xleftarrow{} f$ 
   \For{layer $l$ in $f'$}
     \If{$l$ is not last linear layer}
        \State Freeze parameters in $l$.
     \EndIf
    \State Obtain $\hat{I}_N$ via \cref{eq: weight_neutralization}.
    \State Obtain $U^*$, $S^*$ and $V^*$ via \cref{eq: weighted_svd}.
    \State Replace the final linear layer $l$ with a new layer $\hat{l}_1$ of weights $\hat{I}_N^{-1}U^*_rS^*_r$ and another layer $\hat{l}_2$ of weights $V^*_r$ and $b$.
    \EndFor
    \For{$t$ from 1 to $T$}
        \State Fine-tune $\hat{l}_1$ and $\hat{l}_2$. 
        \EndFor
  \end{algorithmic}
\end{algorithm}

\section{Experiment}
\label{sec: experiment}

In this section, we conduct an empirical analysis of our weight importance neutralization approach. We compare our method with various leading-edge baselines across three real-world datasets. 

\subsection{Experimental Setup}
\textbf{Evaluation Metrics.} We employ two group fairness measures: \textbf{Demographic Parity Distance} ($\Delta_{\text{DP}}$)~\cite{DBLP:flex} and \textbf{Equalized Odds} ($\Delta_{\text{EO}}$)~\cite{equal_opportunity}. $\Delta_{\text{DP}}$ evaluates the absolute difference of the probability of receiving a favorable prediction between the privileged groups $S=1$ and the protected group $S=2$: $\Delta_{\text{DP}}=\vert\mathbb{E}(\hat{Y}=1\mid S=1) - \mathbb{E}(\hat{Y}=1\mid S=2)\vert$. On the other hand, $\Delta_{\text{EO}}$ aims for the sensitive variable $S$ to not influence the prediction of favorable outcomes, conditioned on the ground truth label, which is expressed as $\Delta_{\text{EO}}= \Delta_{\text{TPR}} + \Delta_{\text{FPR}}$, where $\Delta_{\text{TPR}} = |P(\hat{Y}=1 \mid S=1, Y=1) - P(\hat{Y}=1 \mid S=2, Y=1)|$ and $\Delta_{\text{FPR}} = |P(\hat{Y}=1 \mid S=1, Y=-1) - P(\hat{Y}=1 \mid S=2, Y=-1)|$. For both $\Delta_{\text{DP}}$ and $\Delta_{\text{EO}}$, a value closer to 0 indicates minimal bias. For prediction performance, we use the weighted average F1 score, which is calculated by considering the relative portion of instances within different classes. In our experimental results, we report the 100\% - F1 score, which is the error (\textbf{Err}) of the prediction.

\textbf{Benchmark Datasets.} We evaluate the performance of our proposed method using one benchmark tabular dataset and two image datasets. For the tabular dataset, we choose the Adult Income dataset (\textbf{Adult}), and the objective is to determine if an individual's income surpasses \$50K/year~\cite{adultcensus}, with \textit{gender} identified as the sensitive variable. Previous studies indicate a bias in predicting lower earnings for females. For the image datasets, we select the CelebFaces Attributes (\textbf{CelebA})~\cite{celeba} and modified Labeled Faces in the Wild Home (\textbf{LFW+a})~\cite{lfwa_usage}. The CelebA dataset is utilized to discern if the hair in an image is wavy (``WavyHair"), considering gender (``Male") as the sensitive variable where biases have been noted towards males. In the LFW+a dataset~\cite{lfwa_usage}, we augment each image with additional attributes like gender and race (same in CelebA), aiming to classify the identity's gender. The sensitive variable here is ``HeavyMakeup", where literature has shown a strong correlation regarding females. Each dataset is divided into training, validation, and test sets, with the statistics detailed in \cref{tab:dataset_stats}. 

\begin{table}[t]
    \centering
    \begin{tabular}{c c c c }
    \toprule
    Data & Adult & CelebA & LFW+a \\
    \toprule
    \# Training & 12287 & 194599 & 6885 \\
    \# Validation & 5000 & 4000 & 1000 \\
    \# Test & 10315 & 8000 & 5258 \\
    \bottomrule
    \end{tabular}
    \caption{Dataset statistics.}
    \label{tab:dataset_stats}
\end{table}

\textbf{Baselines.} To study the performance of our method, we perform two sets of experiments. The first set of experiments (\cref{sec:exp_first_set}) compares our method with the baseline using SVD-based low-rank decomposition in linear layers during fine-tuning (\textbf{$f$+SVD}), as well as with the traditional transfer learning method that does not use any low-rank decomposition technique (\textbf{TL}). In the second set of experiments (\cref{sec:exp_fair_pretrian} and \cref{sec:exp_fair_after_finetune}), we explore two applications of fairness constraints: one applies these constraints to the pre-training phase to develop a fair pre-trained model, and another incorporates them during fine-tuning. For this set of experiments, we focus on two types of fairness constraints, Equalized Odds (\textbf{EO}) and Demographic Parity (\textbf{DP}). 


\textbf{Implementation Details.} 
For the tabular dataset (Adult), we use a two-layer MLP with ReLU activation as the pre-trained model. Specifically, we split the data with 60\% for training the pre-trained model and the remaining 40\% for a new task. For the two image datasets (LFW+a and CelebA), we adopt ResNet-18~\cite{DBLP:journals/corr/HeZRS15} as the pre-trained model. For all three datasets, we leave out a small validation set for hyperparameter tuning. In our experiments, unless specifically stated, otherwise, we keep the parameters of the pre-trained model frozen and only fine-tune the final linear layer. In order to establish the robustness of our conclusion, the reported results are reported in mean $\pm$ standard deviation format, derived from ten experimental runs.

\begin{figure*}[t]
    \centering
    \includegraphics[width=0.9\linewidth]{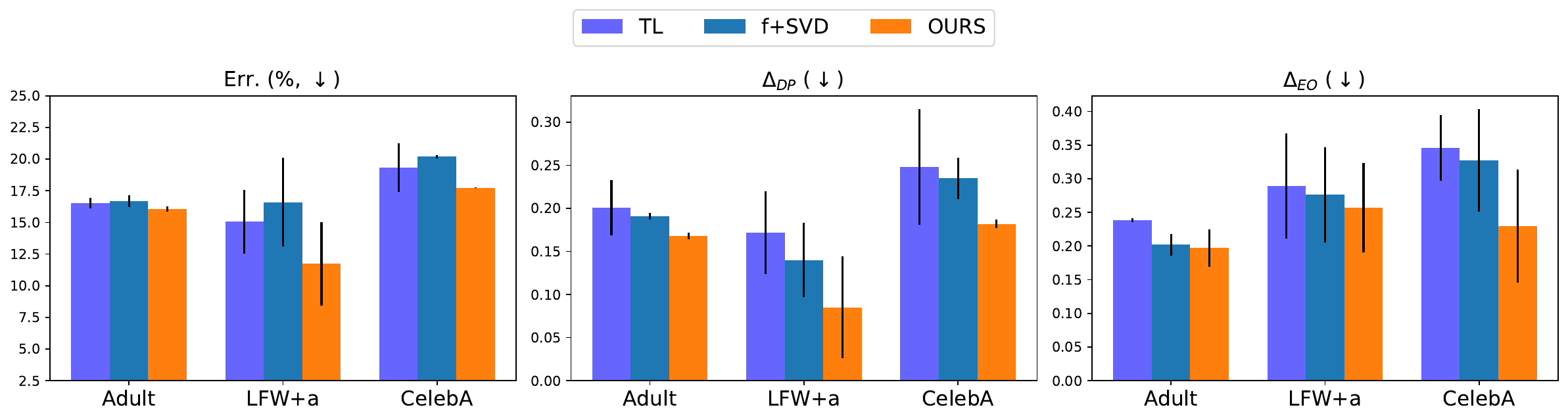}
    \caption{The comparison among TL, $f$+SVD and OURS across three real-world datasets w.r.t. the test errors (F1 score) and fairness violations ($\Delta_{\text{DP}}$ and $\Delta_{\text{EO}}$).}
    \label{fig:compare_results}
\end{figure*}

\begin{table*}[h]
\centering
\begin{tabular}{lccc cc}
\toprule
\multirow{2}{*}{\textbf{Constraints Type}} & \multirow{2}{*}{\textbf{Regularizer Intensity}} & \multicolumn{2}{c}{\textbf{Err (\%$\downarrow$)}} & \multicolumn{2}{c}{\textbf{Bias ($\downarrow$)}} \\
\cmidrule(lr){3-4} \cmidrule(lr){5-6}
& &Pre-train & Fine-tune& Pre-train & Fine-tune\\
\midrule
 & 0.1 & 16.835 & 16.676$_{\pm 0.313}$ & 0.068  & 0.206$_{\pm 0.024}$ (+0.138) \\
\textbf{Equalized Odds} & 0.5 &  17.469 & 17.168$_{\pm 0.941}$ & 0.012 & 0.183$_{\pm 0.037}$ (+0.171) \\
 & 0.9 & 18.744 & 17.343$_{\pm 0.973}$ & 0.008  & 0.186$_{\pm 0.017}$ (+0.178) \\
\midrule
 & 0.1 & 17.332 & 17.479$_{\pm 0.068}$ &  0.125 & 0.171$_{\pm 0.009}$ (+0.046) \\
\textbf{Demographic Parity} & 0.5 & 17.489 & 17.773$_{\pm 0.034}$ & 0.047 & 0.173$_{\pm 0.023}$ (+0.126)\\
 & 0.9 & 20.160 & 17.071$_{\pm 0.036}$ & 0.031 & 0.172$_{\pm 0.018}$ (+0.141) \\
\bottomrule
\end{tabular}
\caption{Experimental results on the Adult dataset using the fair pre-trained model w.r.t. the test errors (F1 score) and fairness violations ($\Delta_{\text{DP}}$ and $\Delta_{\text{EO}}$). 
Parentheses show bias changes from the original fair pre-trained model to the one after fine-tuning, the positive number indicates even with the fair pre-trained model, the fine-tuning process can still introduce bias.}
\label{tab:compare_results_pre_train_fair}
\end{table*}

\subsection{Performance Analysis}
\label{sec:exp_first_set}
In the first experimental setting, we compare three different methods: TL, $f$+SVD, and ours across Adult, LFW+a, and CelebA. We include the complete outcomes of all the baselines w.r.t. both prediction performance and fairness violations, which are provided in \cref{fig:compare_results}. It is evident that our proposed method consistently outperforms the baseline methods across the broad. Our method has the lowest prediction error across all datasets and this superiority also extends to aspects of fairness as well. Specifically, our method significantly reduces both $\Delta_{\text{DP}}$ and $\Delta_{\text{EO}}$, suggesting our method is more effective in mitigating bias compared to baselines. Notably, the performance gap (both prediction and fairness) between our method and baselines is especially apparent for the two image datasets compared to the tabular dataset. 
The observed phenomenon could be attributed to the fact that we trained the pre-trained model on a portion of the tabular dataset. Consequently, the pre-trained model may retain more information relevant to the tabular dataset compared to when it is trained on image datasets. 
The performance of TL and $f$+SVD does not differ too much, suggesting that implementing low-rank decomposition on the final layer does not substantially improve performance. 
Nonetheless, these methods do not sufficiently address bias during fine-tuning, unlike OURS, which is more effective in this regard. These results align with the arguments we put forward in \cref{sec:method_analysis}, which highlight the inherent risk of bias retention in the fine-tuning process. Without deliberate and careful intervention, the fine-tuning process is prone to perpetuating existing discrimination, leading to unfair outcomes.

\subsection{Weight Importance Neutralization with Fair Pre-trained Models}
\label{sec:exp_fair_pretrian}
In this section, we evaluate our approach against the fair pre-trained model $f_F$ on the Adult dataset, focusing on two fairness constraints: (1) Equalized Odds, which require the model to have equal true positive rates and false positive rates across $\mathcal{D}_{S=1}$ and $\mathcal{D}_{S=2}$ and (2) Demographic Parity, which requires the ratio of positive predictions to be identical across two demographic groups. The pre-trained model is developed under these constraints, with varying regularizer intensities from $[0.1, 0.5, 0.9]$. We report the results in \cref{tab:compare_results_pre_train_fair} where the fairness metric corresponds to the applied fairness constraint, e.g., the metric in ``Bias" column indicates $\Delta_{\text{EO}}$ when the Equalized Odds constraint is employed. For both Equalized Odds and Demographic Parity, as the intensity of the regularizer increases, there is a general trend where bias tends to decrease at the expense of increased prediction error. These results highlight the trade-off between accuracy and fairness when incorporating fairness constraints. Specifically, this trend is more obvious for Equalized Odds, however, even with the fairest pre-trained model (the one with $0.9$ regularizer intensity), after fine-tuning, the bias still increases. Therefore, this indicates that even if the pre-trained model is fair, the fine-tuning process can still introduce biases, and this effect is unpredictable. It is important to pay attention to this phenomenon when fine-tuning large models in practice.


\begin{table*}[h]
\centering
\begin{tabular}{lcccc}
\toprule
\multirow{2}{*}{\textbf{Model}} & \multirow{2}{*}{\textbf{Regularizer Intensity}} & \multirow{2}{*}{{\textbf{Err (\%$\downarrow$)}}} & \multicolumn{2}{c}{\textbf{Bias ($\downarrow$)}} \\
\cmidrule(lr){4-5}
& & & $\Delta_{\text{EO}}$ & $\Delta_{\text{DP}}$\\
\midrule
& 0.1 & 16.763$_{\pm 0.223}$ (+0.598) & 0.214$_{\pm 0.027}$ (-0.009) & 0.196$_{\pm 0.027}$ (+0.004)\\
\textbf{Retrain+EO} & 0.5 & 17.440$_{\pm 0.676}$ (+1.275) & 0.150$_{\pm 0.040}$ (-0.073) & 0.183$_{\pm 0.011}$ (-0.009)  \\
& 0.9 & 18.810$_{\pm 0.588}$ (+2.645) & 0.147$_{\pm 0.031}$ (-0.076) & 0.184$_{\pm 0.032}$ (-0.008)\\
\midrule
& 0.1 & 16.752$_{\pm 0.263}$ (+0.587) & 0.206$_{\pm 0.035}$ (-0.017) & 0.174$_{\pm 0.026}$ (-0.018)\\
\textbf{Retrain+DP} & 0.5 & 17.659$_{\pm 0.375}$ (+1.494)  & 0.201$_{\pm 0.024}$ (-0.022) & 0.168$_{\pm 0.023}$ (-0.024) \\
& 0.9 & 18.472$_{\pm 0.228}$ (+2.307) & 0.204$_{\pm 0.023}$ (-0.019) &0.143$_{\pm 0.021}$ (-0.049)\\
\midrule
\textbf{OURS} & --- & 15.829$_{\pm 0.340}$ (-0.336) & 0.197$_{\pm 0.011}$ (-0.026) & 0.180$_{\pm 0.022}$ (-0.012)\\
\bottomrule
\end{tabular}
\caption{Experiment results on the Adult dataset using in-processing fairness methods during fine-tuning w.r.t. the test errors (F1 score) and fairness violations ($\Delta_{\text{DP}}$ and $\Delta_{\text{EO}}$). 
Parentheses show changes from the original pre-trained model to the one after fine-tuning, with negatives indicating improvements. Our method achieves lower $\Delta_{\text{DP}}$ than Retrain+EO and lower $\Delta_{\text{EO}}$ than Retrain+DP. It enhances accuracy and fairness simultaneously.}
\label{tab:compare_results_re_train_fair}
\end{table*}

\subsection{Comparison with In-processing Fairness Methods on New Tasks}
\label{sec:exp_fair_after_finetune}

In this section, a comparative analysis is conducted between our method and those methods implementing fairness constraints during the fine-tuning process. The results are shown in \cref{tab:compare_results_re_train_fair}. Adopting the experimental setting in \cref{sec:exp_fair_pretrian}, we incorporate two types of fairness constraints when fine-tuning on new tasks. The original pre-trained model has a prediction error of $16.165\%$, with fairness violations of $\Delta_{\text{EO}} = 0.223$ and $\Delta_{\text{DP}} = 0.192$. After employing the fairness constraints during fine-tuning, both \textbf{Retrain+EO} and \textbf{Retrain+DP} methods compromise prediction performance to reduce $\Delta_{\text{EO}}$ and $\Delta_{\text{DP}}$. As the regularizer intensity increases, the prediction error increases and the bias decreases, indicating a trade-off between fairness and accuracy. Our method yields comparatively lower prediction errors and biases, while having significantly fewer trainable parameters ($46$ versus $742$, $93.8\%$ reduction), suggesting improved efficiency without the need for adjusting objective functions for new tasks.  Furthermore, our approach provides greater flexibility by eliminating the need to predefine fairness criteria.

\begin{table}[h]
\centering
\begin{tabular}{lccc}
\toprule
\multirow{2}{*}{\textbf{Model}}  & \multirow{2}{*}{{\textbf{Err (\%$\downarrow$)}}} & \multicolumn{2}{c}{\textbf{Bias ($\downarrow$)}} \\
\cmidrule(lr){3-4}
 & & $\Delta_{\text{EO}}$ & $\Delta_{\text{DP}}$\\
 \midrule
 \multicolumn{4}{c}{Dataset: Adult}\\
 \midrule
MLP-3 & 16.678$_{\pm 0.681}$ & 0.185$_{\pm 0.023}$ & 0.162$_{\pm 0.027}$ \\
MLP-4 & 16.237$_{\pm 0.456}$ & 0.173$_{\pm 0.029}$ & 0.170$_{\pm 0.022}$\\
MLP-5 & 16.166$_{\pm 0.337}$ & 0.179$_{\pm 0.012}$ & 0.166$_{\pm 0.014}$\\
\midrule
 \multicolumn{4}{c}{Dataset: LFW+a}\\
\midrule
ResNet-50 & 10.634$_{\pm 0.241}$ & 0.218$_{\pm 0.081}$ & 0.080$_{\pm 0.032}$ \\
DenseNet & 9.775$_{\pm 0.131}$ & 0.232$_{\pm 0.054}$ & 0.073$_{\pm 0.009}$ \\
\midrule
 \multicolumn{4}{c}{Dataset: CelebA}\\
 \midrule
ResNet-50 & 17.582$_{\pm 0.226}$ & 0.243$_{\pm 0.061}$ & 0.182$_{\pm 0.005}$  \\
DenseNet & 18.438$_{\pm 0.174}$ & 0.201$_{\pm 0.028}$ & 0.169$_{\pm 0.009}$ \\
\bottomrule
\end{tabular}
\caption{Ablation study: The test errors (\%) measured using F1 score and fairness violations using different model architectures across different datasets.}
\label{tab:model_structure}
\end{table}

\subsection{Ablation Study}

\begin{figure}[h]
    \centering
    \includegraphics[width=1\linewidth]{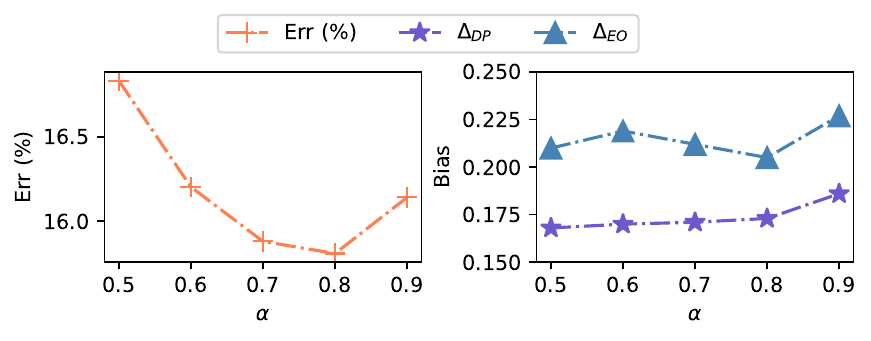}
    \caption{Ablation study: The impact of varying the contribution of different demographic groups in neutralized weight importance on prediction errors and fairness violations using the Adult dataset.}
    \label{fig:fisher_alpha}
\end{figure}

In this section, we conduct an ablation study of our method, which we divide into two parts for discussion. The first part pertains to the model architectures. For the Adult dataset, we experiment with adjusting the number of layers in different MLP structures. For the image datasets, we compare the performance of ResNet-50~\cite{DBLP:journals/corr/HeZRS15} and DenseNet-121~\cite{huang2017densely}. The results are listed in \cref{tab:model_structure}. For the Adult dataset, increasing model complexity leads to marginally improved predictions, with little variation in fairness metrics ($\Delta_{\text{EO}}$ and $\Delta_{\text{DP}}$). For image datasets (LFW+a and CelebA), enhancing model complexity generally improves both predictive performance and fairness, though these improvements are not significant. Overall, the impact of the model's structure on its performance is not obvious. In the second part, we adjust each demographic group's impact as detailed in \cref{eq: weight_neutralization}, diverging from our initial approach which uniformly considers each group's contribution through averaging. Here, we vary the ratio: $\hat{I}_N = \alpha \hat{I}_{S=1} + (1-\alpha)\hat{I}_{S=2}$, $\alpha \in [\frac{1}{2},1)$ and the result is presented in \cref{fig:fisher_alpha}. From the graph, we can see there is a slight decrease in prediction errors, while in the meantime, both $\Delta_{\text{DP}}$ and $\Delta_{\text{EO}}$ generally have an increasing pattern as the value of $\alpha$ increases.


\section{Conclusion}
\label{sec: conclusion}

In this work, we addressed the crucial issue of fairness in fine-tuning. We tackle the limitations of constructing new models from scratch on new tasks, which is computationally intensive in many real-world scenarios. To this end, we proposed a novel
weight importance neutralization strategy during fine-tuning to mitigate bias. Our approach involves assessing the weight importance using Fisher information and then incorporating this into SVD for low-rank approximation. By doing this, our method not only mitigates bias effectively but also enhances the efficiency of fine-tuning large pre-trained models. Our empirical analysis shows the effectiveness of our proposed method and demonstrates that even with a fair pre-trained model, it can still exhibit biases when fine-tuning on new tasks. Future research could further refine this technique, and investigate the applicability of this method across diverse domains with even larger models.

\bibliographystyle{named}
\bibliography{ijcai24}

\begin{thebibliography}{}

\bibitem[\protect\citeauthoryear{Acharya \bgroup \em et al.\egroup }{2018}]{acharya2018online}
Anish Acharya, Rahul Goel, Angeliki Metallinou, and Inderjit Dhillon.
\newblock Online embedding compression for text classification using low rank matrix factorization, 2018.

\bibitem[\protect\citeauthoryear{Agarwal \bgroup \em et al.\egroup }{2018}]{2018_icml_reductions}
Alekh Agarwal, Alina Beygelzimer, Miroslav Dud{\'{\i}}k, John Langford, and Hanna~M. Wallach.
\newblock A reductions approach to fair classification.
\newblock {\em CoRR}, abs/1803.02453, 2018.

\bibitem[\protect\citeauthoryear{Aghajanyan \bgroup \em et al.\egroup }{2021}]{aghajanyan-etal-2021-intrinsic}
Armen Aghajanyan, Sonal Gupta, and Luke Zettlemoyer.
\newblock Intrinsic dimensionality explains the effectiveness of language model fine-tuning.
\newblock In Chengqing Zong, Fei Xia, Wenjie Li, and Roberto Navigli, editors, {\em Proceedings of the 59th Annual Meeting of the Association for Computational Linguistics and the 11th International Joint Conference on Natural Language Processing (Volume 1: Long Papers)}, pages 7319--7328, Online, August 2021. Association for Computational Linguistics.

\bibitem[\protect\citeauthoryear{Ben~Noach and Goldberg}{2020}]{ben-noach-goldberg-2020-compressing}
Matan Ben~Noach and Yoav Goldberg.
\newblock Compressing pre-trained language models by matrix decomposition.
\newblock In Kam-Fai Wong, Kevin Knight, and Hua Wu, editors, {\em Proceedings of the 1st Conference of the Asia-Pacific Chapter of the Association for Computational Linguistics and the 10th International Joint Conference on Natural Language Processing}, pages 884--889, Suzhou, China, December 2020. Association for Computational Linguistics.

\bibitem[\protect\citeauthoryear{{Bilal Zafar} \bgroup \em et al.\egroup }{2015}]{fair_constraints}
Muhammad {Bilal Zafar}, Isabel {Valera}, Manuel {Gomez Rodriguez}, and Krishna~P. {Gummadi}.
\newblock {Fairness Constraints: Mechanisms for Fair Classification}.
\newblock {\em arXiv e-prints}, page arXiv:1507.05259, Jul 2015.

\bibitem[\protect\citeauthoryear{{Bilal Zafar} \bgroup \em et al.\egroup }{2016}]{2016_zafar}
Muhammad {Bilal Zafar}, Isabel {Valera}, Manuel {Gomez Rodriguez}, and Krishna~P. {Gummadi}.
\newblock {Fairness Beyond Disparate Treatment \&amp; Disparate Impact: Learning Classification without Disparate Mistreatment}.
\newblock {\em arXiv e-prints}, page arXiv:1610.08452, Oct 2016.

\bibitem[\protect\citeauthoryear{Bird \bgroup \em et al.\egroup }{2016}]{bird2016exploring}
Sarah Bird, Solon Barocas, Kate Crawford, and Hanna Wallach.
\newblock Exploring or exploiting? social and ethical implications of autonomous experimentation in ai.
\newblock In {\em Workshop on Fairness, Accountability, and Transparency in Machine Learning (FAT-ML), New York University}, page~4, October 2016.

\bibitem[\protect\citeauthoryear{Buolamwini and Gebru}{2018}]{pmlr-v81-buolamwini18a}
Joy Buolamwini and Timnit Gebru.
\newblock Gender shades: Intersectional accuracy disparities in commercial gender classification.
\newblock In Sorelle~A. Friedler and Christo Wilson, editors, {\em Proceedings of the 1st Conference on Fairness, Accountability and Transparency}, volume~81 of {\em Proceedings of Machine Learning Research}, pages 77--91, New York, NY, USA, 23--24 Feb 2018. PMLR.

\bibitem[\protect\citeauthoryear{Calmon \bgroup \em et al.\egroup }{2017}]{NIPS2017_optimised_preprocessing}
Flavio Calmon, Dennis Wei, Bhanukiran Vinzamuri, Karthikeyan Natesan~Ramamurthy, and Kush~R Varshney.
\newblock Optimized pre-processing for discrimination prevention.
\newblock In I.~Guyon, U.~V. Luxburg, S.~Bengio, H.~Wallach, R.~Fergus, S.~Vishwanathan, and R.~Garnett, editors, {\em Advances in Neural Information Processing Systems 30}, pages 3992--4001. Curran Associates, Inc., 2017.

\bibitem[\protect\citeauthoryear{Creager \bgroup \em et al.\egroup }{2019}]{DBLP:flex}
Elliot Creager, David Madras, J{\"{o}}rn{-}Henrik Jacobsen, Marissa~A. Weis, Kevin Swersky, Toniann Pitassi, and Richard~S. Zemel.
\newblock Flexibly fair representation learning by disentanglement.
\newblock {\em CoRR}, abs/1906.02589, 2019.

\bibitem[\protect\citeauthoryear{Dressel and Farid}{2018}]{compas_juve}
Julia Dressel and Hany Farid.
\newblock The accuracy, fairness, and limits of predicting recidivism.
\newblock {\em Science Advances}, 4(1):eaao5580, 2018.

\bibitem[\protect\citeauthoryear{{Dwork} \bgroup \em et al.\egroup }{2011}]{2011_dwork}
Cynthia {Dwork}, Moritz {Hardt}, Toniann {Pitassi}, Omer {Reingold}, and Rich {Zemel}.
\newblock {Fairness Through Awareness}.
\newblock {\em arXiv e-prints}, page arXiv:1104.3913, Apr 2011.

\bibitem[\protect\citeauthoryear{Fu \bgroup \em et al.\egroup }{2023}]{10.1609/aaai.v37i11.26505}
Zihao Fu, Haoran Yang, Anthony Man-Cho So, Wai Lam, Lidong Bing, and Nigel Collier.
\newblock On the effectiveness of parameter-efficient fine-tuning.
\newblock In {\em Proceedings of the Thirty-Seventh AAAI Conference on Artificial Intelligence and Thirty-Fifth Conference on Innovative Applications of Artificial Intelligence and Thirteenth Symposium on Educational Advances in Artificial Intelligence}, AAAI'23/IAAI'23/EAAI'23. AAAI Press, 2023.

\bibitem[\protect\citeauthoryear{Golub and Reinsch}{1971}]{svd}
Gene~H. Golub and Christian Reinsch.
\newblock Singular value decomposition and least squares solutions.
\newblock {\em Linear Algebra}, pages 134--151, 1971.

\bibitem[\protect\citeauthoryear{Hardt \bgroup \em et al.\egroup }{2016}]{equal_opportunity}
Moritz Hardt, Eric Price, and Nathan Srebro.
\newblock Equality of opportunity in supervised learning.
\newblock {\em CoRR}, abs/1610.02413, 2016.

\bibitem[\protect\citeauthoryear{He \bgroup \em et al.\egroup }{2016}]{DBLP:journals/corr/HeZRS15}
Kaiming He, Xiangyu Zhang, Shaoqing Ren, and Jian Sun.
\newblock Deep residual learning for image recognition.
\newblock In {\em 2016 {IEEE} Conference on Computer Vision and Pattern Recognition, {CVPR} 2016, Las Vegas, NV, USA, June 27-30, 2016}, pages 770--778. {IEEE} Computer Society, 2016.

\bibitem[\protect\citeauthoryear{He \bgroup \em et al.\egroup }{2022}]{he2022towards}
Junxian He, Chunting Zhou, Xuezhe Ma, Taylor Berg-Kirkpatrick, and Graham Neubig.
\newblock Towards a unified view of parameter-efficient transfer learning.
\newblock In {\em International Conference on Learning Representations}, 2022.

\bibitem[\protect\citeauthoryear{Hsu \bgroup \em et al.\egroup }{2022}]{hsu2022language}
Yen-Chang Hsu, Ting Hua, Sungen Chang, Qian Lou, Yilin Shen, and Hongxia Jin.
\newblock Language model compression with weighted low-rank factorization.
\newblock In {\em International Conference on Learning Representations}, 2022.

\bibitem[\protect\citeauthoryear{Hu \bgroup \em et al.\egroup }{2022}]{hu2022lora}
Edward~J Hu, yelong shen, Phillip Wallis, Zeyuan Allen-Zhu, Yuanzhi Li, Shean Wang, Lu~Wang, and Weizhu Chen.
\newblock Lo{RA}: Low-rank adaptation of large language models.
\newblock In {\em International Conference on Learning Representations}, 2022.

\bibitem[\protect\citeauthoryear{Huang \bgroup \em et al.\egroup }{2017}]{huang2017densely}
Gao Huang, Zhuang Liu, Laurens van~der Maaten, and Kilian~Q. Weinberger.
\newblock Densely connected convolutional networks.
\newblock In {\em CVPR}, pages 2261--2269. IEEE Computer Society, 2017.

\bibitem[\protect\citeauthoryear{Iosifidis and Ntoutsi}{2019}]{adafair}
Vasileios Iosifidis and Eirini Ntoutsi.
\newblock Adafair: Cumulative fairness adaptive boosting.
\newblock In {\em Proceedings of the 28th ACM International Conference on Information and Knowledge Management}, CIKM '19, page 781–790, New York, NY, USA, 2019. Association for Computing Machinery.

\bibitem[\protect\citeauthoryear{Jaderberg \bgroup \em et al.\egroup }{2014}]{jaderberg2014speeding}
Max Jaderberg, Andrea Vedaldi, and Andrew Zisserman.
\newblock Speeding up convolutional neural networks with low rank expansions, 2014.

\bibitem[\protect\citeauthoryear{Kamishima \bgroup \em et al.\egroup }{2012}]{Kamishima_2011}
Toshihiro Kamishima, Shotaro Akaho, Hideki Asoh, and Jun Sakuma.
\newblock Fairness-aware classifier with prejudice remover regularizer.
\newblock In Peter~A. Flach, Tijl De~Bie, and Nello Cristianini, editors, {\em Machine Learning and Knowledge Discovery in Databases}, pages 35--50, Berlin, Heidelberg, 2012. Springer Berlin Heidelberg.

\bibitem[\protect\citeauthoryear{Khandani \bgroup \em et al.\egroup }{2010}]{credit_example}
Amir~E. Khandani, Adlar~J. Kim, and Andrew~W. Lo.
\newblock Consumer credit-risk models via machine-learning algorithms.
\newblock {\em Journal of Banking \& Finance}, 34(11):2767 -- 2787, 2010.

\bibitem[\protect\citeauthoryear{Kim \bgroup \em et al.\egroup }{2015}]{medical}
Sung-Eun Kim, Hee~Young Paik, Hyuk Yoon, Jung Lee, Nayoung Kim, and Mi-Kyung Sung.
\newblock Sex- and gender-specific disparities in colorectal cancer risk.
\newblock {\em World journal of gastroenterology : WJG}, 21:5167--5175, 05 2015.

\bibitem[\protect\citeauthoryear{Kohavi}{1996}]{adultcensus}
Ron Kohavi.
\newblock Scaling up the accuracy of naive-bayes classifiers: A decision-tree hybrid.
\newblock In {\em Proceedings of the Second International Conference on Knowledge Discovery and Data Mining}, KDD'96, page 202–207. AAAI Press, 1996.

\bibitem[\protect\citeauthoryear{Lin \bgroup \em et al.\egroup }{2020}]{lin-etal-2020-exploring}
Zhaojiang Lin, Andrea Madotto, and Pascale Fung.
\newblock Exploring versatile generative language model via parameter-efficient transfer learning.
\newblock In Trevor Cohn, Yulan He, and Yang Liu, editors, {\em Findings of the Association for Computational Linguistics: EMNLP 2020}, pages 441--459, Online, November 2020. Association for Computational Linguistics.

\bibitem[\protect\citeauthoryear{Liu \bgroup \em et al.\egroup }{2015}]{celeba}
Ziwei Liu, Ping Luo, Xiaogang Wang, and Xiaoou Tang.
\newblock Deep learning face attributes in the wild.
\newblock In {\em 2015 IEEE International Conference on Computer Vision (ICCV)}, pages 3730--3738, 2015.

\bibitem[\protect\citeauthoryear{Liu \bgroup \em et al.\egroup }{2019}]{liu2019roberta}
Yinhan Liu, Myle Ott, Naman Goyal, Jingfei Du, Mandar Joshi, Danqi Chen, Omer Levy, Mike Lewis, Luke Zettlemoyer, and Veselin Stoyanov.
\newblock Roberta: A robustly optimized bert pretraining approach, 2019.
\newblock cite arxiv:1907.11692.

\bibitem[\protect\citeauthoryear{Oneto \bgroup \em et al.\egroup }{2020}]{NEURIPS2020_mmd}
Luca Oneto, Michele Donini, Giulia Luise, Carlo Ciliberto, Andreas Maurer, and Massimiliano Pontil.
\newblock Exploiting mmd and sinkhorn divergences for fair and transferable representation learning.
\newblock In H.~Larochelle, M.~Ranzato, R.~Hadsell, M.F. Balcan, and H.~Lin, editors, {\em Advances in Neural Information Processing Systems}, volume~33, pages 15360--15370. Curran Associates, Inc., 2020.

\bibitem[\protect\citeauthoryear{Pascanu and Bengio}{2014}]{fisher_info}
Razvan Pascanu and Yoshua Bengio.
\newblock Revisiting natural gradient for deep networks.
\newblock In Yoshua Bengio and Yann LeCun, editors, {\em 2nd International Conference on Learning Representations, {ICLR} 2014, Banff, AB, Canada, April 14-16, 2014, Conference Track Proceedings}, 2014.

\bibitem[\protect\citeauthoryear{Roh \bgroup \em et al.\egroup }{2020}]{pmlr-v119-roh20a}
Yuji Roh, Kangwook Lee, Steven Whang, and Changho Suh.
\newblock {FR}-train: A mutual information-based approach to fair and robust training.
\newblock In Hal~Daumé III and Aarti Singh, editors, {\em Proceedings of the 37th International Conference on Machine Learning}, volume 119 of {\em Proceedings of Machine Learning Research}, pages 8147--8157. PMLR, 13--18 Jul 2020.

\bibitem[\protect\citeauthoryear{Tayaranian~Hosseini \bgroup \em et al.\egroup }{2023}]{tayaranian-hosseini-etal-2023-towards}
Mohammadreza Tayaranian~Hosseini, Alireza Ghaffari, Marzieh~S. Tahaei, Mehdi Rezagholizadeh, Masoud Asgharian, and Vahid Partovi~Nia.
\newblock Towards fine-tuning pre-trained language models with integer forward and backward propagation.
\newblock In Andreas Vlachos and Isabelle Augenstein, editors, {\em Findings of the Association for Computational Linguistics: EACL 2023}, pages 1912--1921, Dubrovnik, Croatia, May 2023. Association for Computational Linguistics.

\bibitem[\protect\citeauthoryear{Wolf \bgroup \em et al.\egroup }{2011}]{lfwa_usage}
Lior Wolf, Tal Hassner, and Yaniv Taigman.
\newblock {Effective Unconstrained Face Recognition by Combining Multiple Descriptors and Learned Background Statistics}.
\newblock {\em IEEE Transactions on Pattern Analysis and Machine Intelligence}, 33(10):1978--1990, 2011.

\bibitem[\protect\citeauthoryear{Xu \bgroup \em et al.\egroup }{2021}]{xu-etal-2021-raise}
Runxin Xu, Fuli Luo, Zhiyuan Zhang, Chuanqi Tan, Baobao Chang, Songfang Huang, and Fei Huang.
\newblock Raise a child in large language model: Towards effective and generalizable fine-tuning.
\newblock In Marie-Francine Moens, Xuanjing Huang, Lucia Specia, and Scott Wen-tau Yih, editors, {\em Proceedings of the 2021 Conference on Empirical Methods in Natural Language Processing}, pages 9514--9528, Online and Punta Cana, Dominican Republic, November 2021. Association for Computational Linguistics.

\bibitem[\protect\citeauthoryear{Zemel \bgroup \em et al.\egroup }{2013}]{icml_2013}
Rich Zemel, Yu~Wu, Kevin Swersky, Toni Pitassi, and Cynthia Dwork.
\newblock Learning fair representations.
\newblock In Sanjoy Dasgupta and David McAllester, editors, {\em Proceedings of the 30th International Conference on Machine Learning}, volume~28 of {\em Proceedings of Machine Learning Research}, pages 325--333, Atlanta, Georgia, USA, 17--19 Jun 2013. PMLR.

\bibitem[\protect\citeauthoryear{Zhang \bgroup \em et al.\egroup }{2018}]{zhang_adversarial}
Brian~Hu Zhang, Blake Lemoine, and Margaret Mitchell.
\newblock Mitigating unwanted biases with adversarial learning.
\newblock In {\em Proceedings of the 2018 AAAI/ACM Conference on AI, Ethics, and Society}, AIES '18, page 335–340, New York, NY, USA, 2018. Association for Computing Machinery.

\bibitem[\protect\citeauthoryear{Zhang \bgroup \em et al.\egroup }{2023}]{pmlr-v206-zhang23g}
Yixuan Zhang, Feng Zhou, Zhidong Li, Yang Wang, and Fang Chen.
\newblock Fair representation learning with unreliable labels.
\newblock In Francisco Ruiz, Jennifer Dy, and Jan-Willem van~de Meent, editors, {\em Proceedings of The 26th International Conference on Artificial Intelligence and Statistics}, volume 206 of {\em Proceedings of Machine Learning Research}, pages 4655--4667. PMLR, 25--27 Apr 2023.

\end{thebibliography}

\end{document}